\documentclass{article}
\usepackage{float}
\usepackage[preprint]{corl_2026}

\usepackage{amssymb}
\usepackage{booktabs}
\usepackage{graphicx}
\usepackage{appendix}
\usepackage{multirow}
\usepackage{wrapfig}
\usepackage{needspace}
\usepackage{capt-of}
\usepackage{amsmath}
\usepackage[skins]{tcolorbox}

\definecolor{resultgreen}{RGB}{0,128,0}
\definecolor{mydarkblue}{rgb}{0,0.08,0.45}
\definecolor{AtkRed}{RGB}{239,50,36}
\definecolor{BenGreen}{RGB}{90,201,59}
\definecolor{FeatMag}{RGB}{192,67,128}
\definecolor{DetBlue}{RGB}{31,111,191}
\colorlet{AtkRedText}{AtkRed!85!black}
\colorlet{BenGreenText}{BenGreen!62!black}
\colorlet{StepGray}{black!62}

\newcommand{\atk}[1]{\textcolor{AtkRedText}{#1}}
\newcommand{\ben}[1]{\textcolor{BenGreenText}{#1}}
\DeclareRobustCommand{\stepbadgebox}[2]{\tikz[baseline=(b.base)]\node[circle,fill=#1,text=white,inner sep=0pt,minimum size=1.05em,font=\scriptsize\bfseries\sffamily](b){#2};}
\newcommand{\stepbadge}[2]{\texorpdfstring{\stepbadgebox{#1}{#2}}{(#2)}}
\newcommand{\gate}{\textcolor{DetBlue}{g_t}}

\title{Detect and Suppress: A Mechanistic Defense against Adversarial Patches in VLA Models}

\author{
  Yukiya Horiba \\
  Keio University \\
  \And
  Koshiro Aoki \\
  Waseda University \\
  \And
  Shunsuke Yasuki \\
  The University of Tokyo \\
  \And
  Bum Jun Kim \\
  The University of Tokyo \\
  \And
  Taiki Miyanishi \\
  The University of Tokyo \\
}

\begin{document}
\maketitle

\begin{figure}[h]
    \centering
    \includegraphics[width=0.9\linewidth]{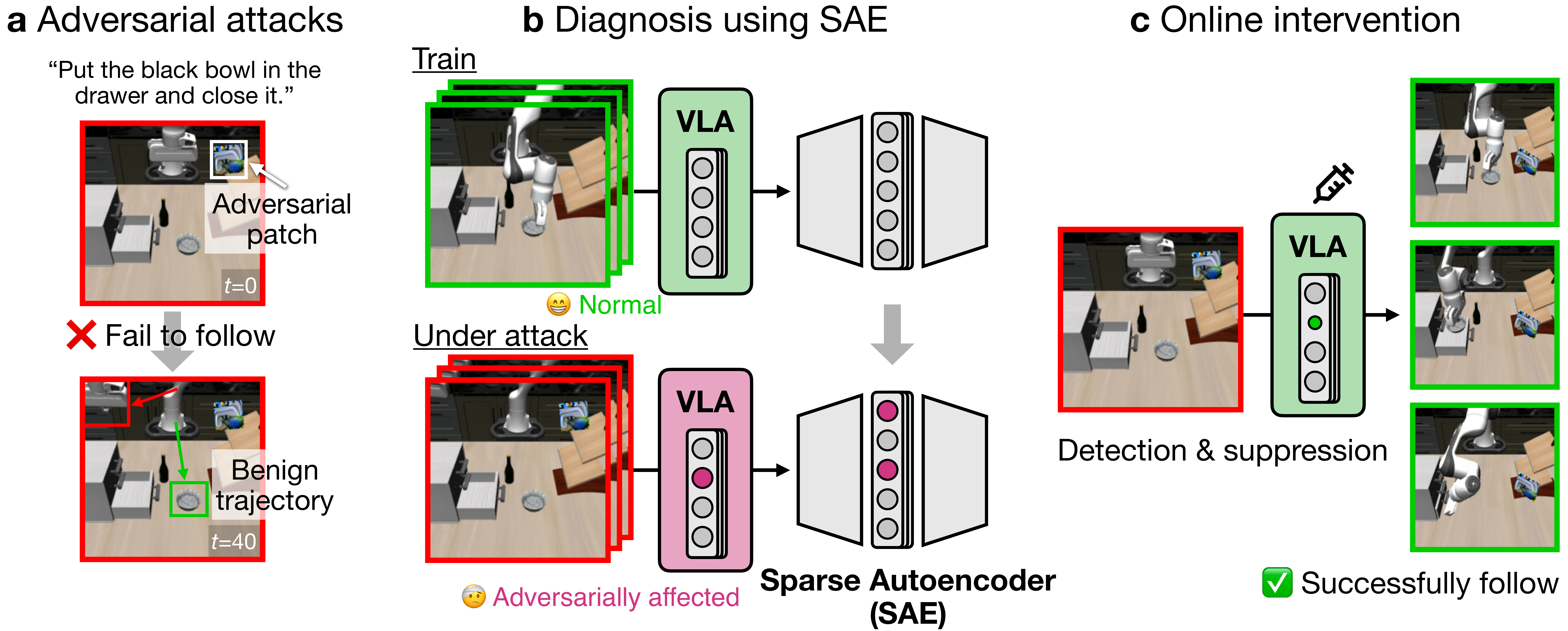}
    \caption{{Overview of our SAE-based defense.}}
    \label{fig:placeholder}
\end{figure}

\begin{abstract}
    Adversarial patches can disrupt Vision-Language-Action (VLA) models by manipulating visual observations, leading to failures in robot control.
    However, it remains poorly understood which internal mechanisms underlie these failures and how targeted interventions can mitigate them.
    In this work, we mechanistically analyze VLA representations using a sparse autoencoder (SAE) and identify a feature whose activation strongly correlates with the presence of an adversarial patch.
    Based on this analysis, we suppress the identified feature at inference time only when a linear probe detects an attack.
    This intervention improves robustness without the cost of fine-tuning the VLA.
    We evaluate our method against VLA adversarial patch attacks on LIBERO-10.
    Conditional intervention improves success rate under intermittent attacks, whereas continuously applying the same intervention substantially degrades policy performance.
    These results show that attack-related internal representations can provide useful targets for VLA adversarial defense and that controlling when to intervene is important for limiting disruption to nominal policy behavior.
\end{abstract}


\section{Introduction}
\label{sec:introduction}

Adversarial patches can disrupt vision-language-action (VLA) policies by altering the visual observations used for robot control~\cite{wang2025exploring,xu2025model}.
Existing VLA defenses modify either the inputs, through image preprocessing~\cite{wang2026partially}, or the model parameters, through adversarial fine-tuning~\cite{xu2025model}.
Adversarial fine-tuning, however, incurs the computational cost of additional policy training.
We ask whether VLA robustness can instead be improved by directly targeting attack-related internal features at inference time, without fine-tuning the policy.

Mechanistic interpretability provides a basis for investigating this question.
Recent studies have identified internal directions and sparse autoencoder (SAE) features whose manipulation influences VLA behavior~\cite{haon2025mechanistic,swann2026sparse,jin2026event,buurmeijer2026observing}.
These studies primarily examine nominal behavior, leaving the use of such features for adversarial defense underexplored.
However, indiscriminately suppressing attack-related features may also disrupt nominal policy behavior.
An effective defense therefore requires examining both which feature to target and when to suppress it.

We investigate these questions through SAE-based analysis and conditional intervention in VLA representations.
We identify an SAE feature associated with patch presence and train a linear probe to detect attacks from the VLA's internal state.
At inference time, we suppress the feature's contribution only when the probe detects an attack, while keeping the VLA parameters fixed.
On LIBERO-10~\cite{liu2023libero} under Untargeted Action Discrepancy Attack (UADA)~\cite{wang2025exploring}, this conditional intervention improves task success under intermittent attacks, whereas continuous suppression of the same feature substantially degrades performance.
Our results suggest that internal features associated with adversarial patches can serve as targets for VLA defense, and that intervening only during detected attacks limits disruption of nominal behavior.

\begin{wrapfigure}{r}{0.45\textwidth}
    \vspace{-4.2em}
    \centering
    \includegraphics[width=\linewidth]{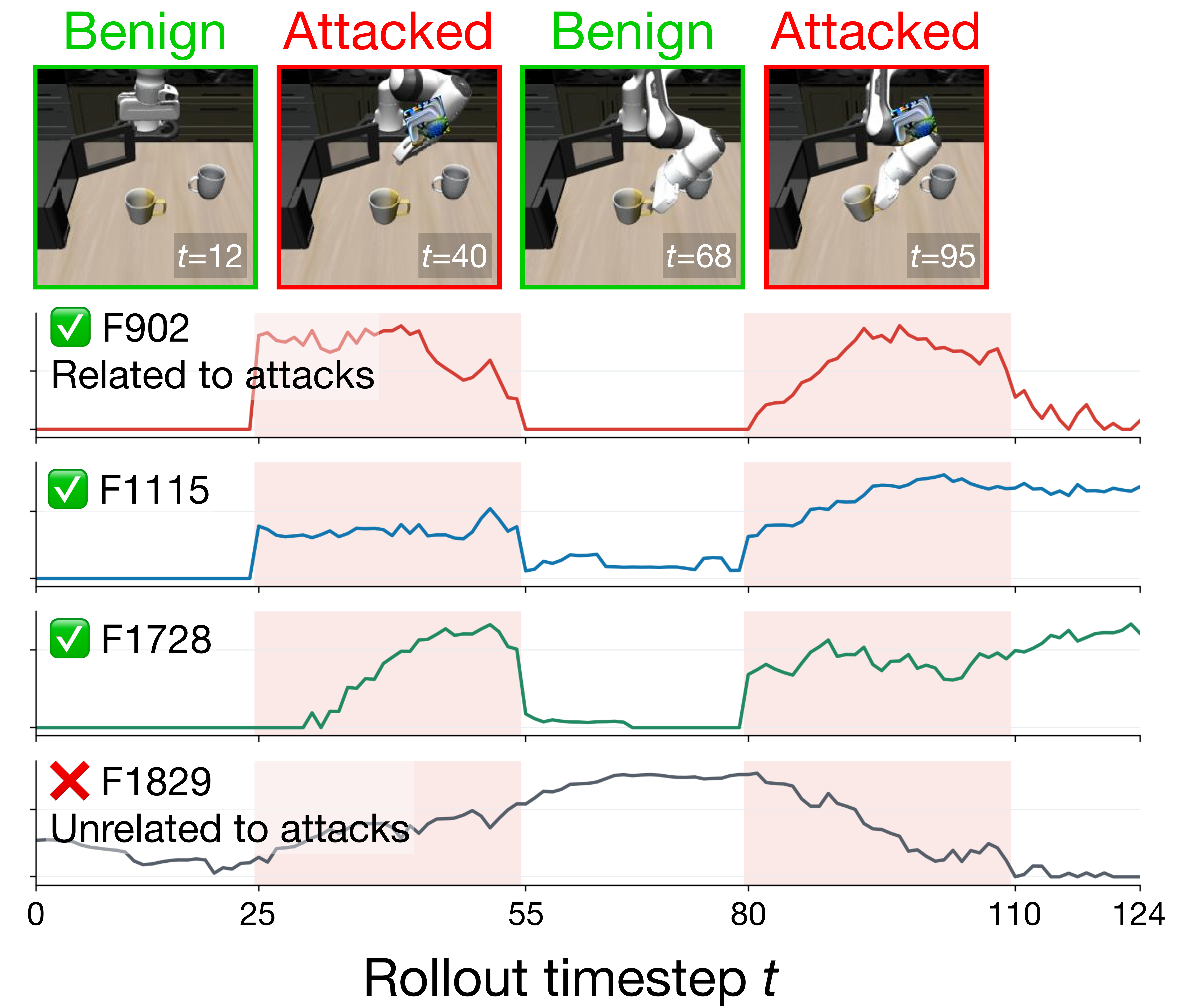}
    \vspace{-1.5em}
    \captionof{figure}{SAE features associated with adversarial patch presence.}
    \label{fig:motivation}
\end{wrapfigure}

\section{Motivation: Attack-Related Features}
\label{sec:motivating_analysis}

Before introducing our method, we examine how adversarial patches are reflected in VLA representations.
We train an SAE on internal representations collected from clean trajectories on LIBERO tasks.
We then repeatedly switch an adversarial patch generated by UADA~\cite{wang2025exploring} on and off during a rollout in order to examine how SAE feature activations respond to patch presence.
Figure~\ref{fig:motivation} shows that several SAE features track patch presence over time, a pattern also observed across multiple episodes.
These observations motivate us to investigate whether suppressing these features only when an attack is detected can improve policy robustness.

\par
\vspace{0.5em}
\noindent
\begin{minipage}{\textwidth}
\section{Methodology: Conditional Feature Intervention}
\label{sec:methodlogy}
\end{minipage}

\paragraph{Problem Setup.}
Our goal is to improve task success under adversarial patch attacks while limiting performance degradation under clean observations.
A VLA policy generates an action \(a_t = \pi(o_t, l)\) from a visual observation \(o_t\) and a language instruction \(l\).
The observation contains agent-view images, wrist-view images, or both, depending on the model.
We consider UADA~\cite{wang2025exploring}, an untargeted attack that optimizes an adversarial patch \(\delta\) to make predicted actions deviate from the ground-truth actions.
Applying the patch yields an attacked observation \(o_t^{\mathrm{adv}} = \mathcal{A}(o_t, \delta)\), while leaving \(l\) unchanged.

\paragraph{Overview.}
We use a TopK SAE~\cite{gao2025scaling} to analyze how adversarial patches affect internal VLA representations.
Our method suppresses a selected attack-related feature only when an attack is detected, aiming to improve robustness while limiting performance degradation under benign conditions.
The method consists of three components: \stepbadge{FeatMag}{1}~identifying an attack-related feature using an SAE, \stepbadge{DetBlue}{2}~detecting adversarial patches with a linear probe, and \stepbadge{StepGray}{3}~conditionally suppressing the selected feature based on the detection result.

\paragraph{\stepbadge{FeatMag}{1}~Attack-Related Feature Identification.}
Let \(h_{t,r}^{(i)} \in \mathbb{R}^{d}\) denote the representation of the \(r\)-th prefix token at layer \(i\) and control step \(t\), where \(r\in\{1,\ldots,N_t\}\) and \(N_t\) is the number of prefix tokens.
We mean-pool the prefix-token representations as
\(\bar{h}_t^{(i)} = \frac{1}{N_t}\sum_{r=1}^{N_t} h_{t,r}^{(i)}\)
and obtain SAE feature activations \(z_t^{(i)} = E^{(i)}(\bar{h}_t^{(i)})\).
Using rollouts in which the adversarial patch is switched on and off, we measure the correlation between SAE feature activations and patch presence across candidate layers.
We select the most consistently correlated layer-feature pair \(\textcolor{FeatMag}{(\ell^\star,j^\star)}\) as the intervention target and denote its decoder direction by \(\textcolor{FeatMag}{d_{j^\star}^{(\ell^\star)}}\).

\paragraph{\stepbadge{DetBlue}{2}~Attack Detection.}
Continuously suppressing the selected feature can degrade policy performance under clean observations.
We therefore train a linear probe to detect patch presence from the token-mean-pooled VLA representation \(\bar{h}_t^{(k)}\) at layer \(k\), where \(k \le \textcolor{FeatMag}{\ell^\star}\) so that the detection result is available before intervention.
The probe estimates the attack probability as \(p_t = \sigma(w^\top \bar{h}_t^{(k)} + b)\), where \(\sigma\) is the sigmoid function, and \(w\) and \(b\) are learned parameters.
We define the intervention gate as \(\gate = \mathbb{I}[p_t > \tau]\), where \(\mathbb{I}\) is the indicator function and \(\tau\) is the detection threshold.

\paragraph{\stepbadge{StepGray}{3}~Conditional Feature Intervention.}
At inference time, we compute the selected feature activation from the token-mean-pooled representation and suppress its contribution in each prefix-token representation at layer \(\ell^\star\):
\begin{equation}
    \widetilde{h}_{t,r}^{(\ell^\star)}
    =
    h_{t,r}^{(\ell^\star)}
    \textcolor{FeatMag}{-}
    \underbrace{\gate}_{\textcolor{DetBlue}{\textsf{detect}}}
    \, \lambda\,
    \underbrace{
        \textcolor{FeatMag}{
            z_{t,j^\star}^{(\ell^\star)}
            \, d_{j^\star}^{(\ell^\star)}
        }
    }_{\textcolor{FeatMag}{\textsf{suppress}}},
    \label{eq:intervention}
\end{equation}
where \(\lambda\) is the intervention strength.
When \(0 < \lambda < 1\), the intervention attenuates the selected feature contribution, while \(\lambda=1\) removes it completely.
For \(\lambda>1\), the intervention extends beyond feature removal and steers the representation in the opposite direction along the corresponding decoder direction.
The intervention is applied to all prefix tokens only when the probe detects an attack (\(g_t=1\)); otherwise, the representations remain unchanged.

\section{Experiments}
\label{sec:result}

\subsection{Experimental Setup}

\paragraph{Models and Benchmark.}
We evaluate $\pi_{0.5}$~\cite{black2025pi05} and SmolVLA~\cite{shukor2025smolvla} on LIBERO-10~\cite{liu2023libero}.
For $\pi_{0.5}$, we use a model that uses only the agent camera, while SmolVLA uses both agent and wrist cameras.
Each main evaluation setting is evaluated on 50 initial states for each of the 10 tasks, for a total of 500 rollouts.

\paragraph{Attack Setting.}
We use UADA~\cite{wang2025exploring} targeting degree of freedom (DoF) 1 and generate one universal adversarial patch for each victim model.
Our main evaluation considers intermittent attacks in which the patch appears $n=3$ or $n=10$ times during each rollout.

\paragraph{Defense and Comparisons.}
For each victim model, we identify an attack-related SAE feature and train a logistic regression probe to detect patch presence from internal representations.
At inference time, our method suppresses the selected feature only when the probe detects the patch.
We compare against the attacked policy without intervention (\textit{Baseline}), continuous intervention without detection (\textit{w/o Det.}), and intervention applied only when the patch is actually present (\textit{w/ Oracle Det.}).
The probe threshold is set on separate construction data to keep the false positive rate below $2\%$.

\paragraph{Evaluation.}
We report task success rate (SR) as the primary metric.
The selected SAE feature, probe, and intervention strength are fixed before evaluation.
Details of SAE training, feature selection, probe training, and intervention settings are provided in Appendix~\ref{sec:experimental_details}.

\subsection{Main Results}

\begin{wraptable}{r}{0.40\textwidth}
    \vspace{-3.5em}
    \centering

    \caption{Detection performance (\%).}
    \label{tab:probe_detection}

    \small
    \begin{tabular}{lccc}
        \toprule
        Setting & Recall & FPR  & Precision \\
        \midrule
        $n=3$   & 81.76  & 0.39 & 92.71     \\
        $n=10$  & 84.91  & 0.71 & 96.45     \\
        \bottomrule
    \end{tabular}

    \vspace{-1em}
\end{wraptable}

\paragraph{Probe Detection Performance.}
We first evaluate whether the probe detects attacks during LIBERO-10 rollouts while rarely raising false positives, which would trigger unnecessary suppression.
The probe uses representations from layer 5 of the $\pi_{0.5}$ LLM backbone.
As shown in Table~\ref{tab:probe_detection}, it achieves recall above $81\%$ and precision above $92\%$ in both settings, with a false positive rate (FPR) of $0.39\%$ for $n=3$ and $0.71\%$ for $n=10$.
These FPRs remain below the $2\%$ target used for threshold selection on separate construction data.
These results support using the probe for conditional feature suppression.

\begin{wraptable}{r}{0.5\textwidth}
    \vspace{-1.8em}
    \centering
    \caption{SR (\%) under \atk{attack} conditions.}
    \label{tab:main_results}

    \small
    \setlength{\tabcolsep}{4pt}
    \renewcommand{\arraystretch}{0.88}

    \resizebox{\linewidth}{!}{%
        \begin{tabular}{lcccc}
            \toprule
            \multirow{2}{*}{Method}
             & \multicolumn{2}{c}{\textbf{Victim: $\pi_{0.5}$}}
             & \multicolumn{2}{c}{\textbf{Victim:} SmolVLA}                     \\
            \cmidrule(lr){2-3}
            \cmidrule(lr){4-5}
             & $n=3$                                            & $n=10$
             & $n=3$                                            & $n=10$        \\
            \midrule

            Baseline
             & 46.8                                             & 26.0
             & 18.6                                             & 6.2           \\

            \textbf{Ours}
             & \textbf{49.2}                                    & \textbf{32.4}
             & \textbf{19.6}                                    & \textbf{7.6}  \\

            \midrule

            w/o Det.
             & 9.4                                              & 16.2
             & 18.4                                             & 4.8           \\

            w/ Oracle Det.
             & 52.2                                             & 29.2
             & 22.8                                             & 7.4           \\

            \bottomrule
        \end{tabular}%
    }

    \vspace{-0.5em}
\end{wraptable}

\paragraph{Effect of Conditional Intervention.}
Table~\ref{tab:main_results} shows that increasing patch exposure from $n=3$ to $n=10$ substantially reduces baseline SR for both $\pi_{0.5}$ and SmolVLA.
Our conditional intervention improves SR across all four settings, with the largest gain from $26.0\%$ to $32.4\%$ for $\pi_{0.5}$ at $n=10$.
In contrast, continuous intervention (\textit{w/o Det.}) can severely degrade performance, particularly for $\pi_{0.5}$.
Our method also achieves SR comparable to intervention based on ground-truth patch presence (\textit{w/ Oracle Det.}).
These results indicate that SAE feature suppression can improve robustness, but its effectiveness depends on when the intervention is applied.

\begin{wraptable}{r}{0.44\textwidth}
    \vspace{-1.8em}
    \centering
    \caption{SR (\%) under \ben{benign} conditions.}
    \label{tab:benign_results}
    \small
    \setlength{\tabcolsep}{7pt}
    \begin{tabular}{lcc}
        \toprule
        Method
         & $\pi_{0.5}$
         & SmolVLA                                                   \\
        \midrule
        Baseline
         & 54.4
         & 41.4                                                      \\
        \textbf{Ours}
         & 54.6 {\scriptsize\textcolor{resultgreen}{$\uparrow 0.2$}}
         & 36.6 {\scriptsize\textcolor{red}{$\downarrow 4.8$}}       \\
        w/o Det.
         & 35.2 {\scriptsize\textcolor{red}{$\downarrow 19.2$}}
         & 19.0 {\scriptsize\textcolor{red}{$\downarrow 22.4$}}      \\
        \bottomrule
    \end{tabular}
    \vspace{-0.5em}
\end{wraptable}

\paragraph{Performance under Benign Conditions.}
We evaluate performance under benign conditions to examine whether the defense disrupts task execution in the absence of attacks.
As shown in Table~\ref{tab:benign_results}, our conditional intervention maintains baseline performance on $\pi_{0.5}$ ($54.6\%$ vs.\ $54.4\%$), while SR decreases from $41.4\%$ to $36.6\%$ on SmolVLA.
In contrast, continuous intervention (\textit{w/o Det.}) substantially reduces SR to $35.2\%$ and $19.0\%$, respectively.
These results indicate that restricting suppression to detected attacks is important for preventing performance degradation under benign conditions.

\begin{wrapfigure}{r}{0.5\textwidth}
    \vspace{-1.8em}
    \centering
    \includegraphics[width=\linewidth]{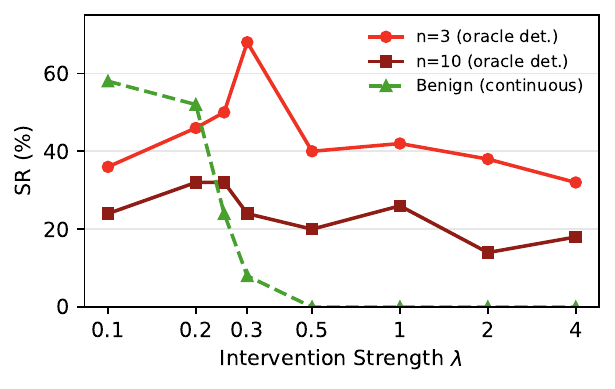}
    \caption{Effect of intervention strength $\lambda$ on SR under \atk{attack} and \ben{benign} conditions.}
    \label{fig:intervention_strength}

    \vspace{-0.5em}
\end{wrapfigure}

\paragraph{Effect of Intervention Strength.}
We examine how the intervention strength $\lambda$ affects performance on $\pi_{0.5}$.
For this analysis, separate from the main evaluation, we use 10 closed-loop episodes per seed across five random seeds (50 rollouts total).
To isolate this effect from detection errors, we use oracle detection under attack and continuous intervention under benign conditions.
As shown in Fig.~\ref{fig:intervention_strength}, under attack, SR peaks around $\lambda=0.3$ for $n=3$ and between $\lambda=0.2$ and $0.25$ for $n=10$, then declines with stronger intervention.
Under benign conditions, continuous intervention sharply reduces SR as $\lambda$ increases.
These results reveal a trade-off between robustness under attack and benign policy performance.

\section{Conclusion}
We presented an SAE-based defense that suppresses an attack-related feature only when a linear probe detects an attack, without additional fine-tuning of the VLA policy.
On LIBERO-10, conditional suppression improves task success under intermittent UADA attacks on $\pi_{0.5}$ and SmolVLA, while causing less performance degradation under benign conditions than continuous suppression.
Our results highlight the importance of selecting attack-related features and controlling when and how strongly to suppress them.
On SmolVLA, however, the gains are modest, and benign performance still degrades.
Future work should evaluate robustness to unseen attacks and test the defense on physical robots.

\clearpage
\acknowledgments{This work was supported by JST PRESTO (Grant Number JPMJPR22P8) and JSPS KAKENHI (Grant Number 25K03177), Japan.}

\bibliography{main}

\clearpage
\appendix

\section{Related Work}
\label{sec:related_work}

\subsection{Adversarial Patch Attacks and Defenses.}
Adversarial patches are localized visual patterns that alter the predictions of image classifiers and object detectors~\cite{brown2017adversarial,liu2018dpatch}.
Prior work has made such perturbations robust to changes in viewing conditions, including physically realizable ones~\cite{eykholt2018robust,athalye2018synthesizing}.
Defenses include masking corrupted features~\cite{xiang2021patchguard}, certified input masking~\cite{xiang2022patchcleanser}, and patch localization and removal~\cite{jing2024pad}.

In VLA models, visual perturbations can affect successive actions and disrupt task execution.
Wang et al.~\cite{wang2025exploring} introduced patch attacks that either increase action discrepancies, as in UADA, or manipulate robot trajectories.
Subsequent studies attack visual embeddings and image--instruction alignment~\cite{xu2025model}, design attacks spanning textual, visual, and cross-modal perturbations~\cite{yan2025alignment}, craft universal patches that transfer across models, tasks, and viewpoints~\cite{lu2026robots}, and optimize patches using only a short prefix of an execution trajectory~\cite{wang2026partially}.

Existing VLA defenses either preprocess the input images or fine-tune the model.
JPEG compression and Gaussian blur provide limited protection against the patch attacks studied in~\cite{wang2026partially}.
Xu et al.~\cite{xu2025model} adversarially fine-tune the visual encoder to align the representations of clean and perturbed inputs.
Our method instead intervenes directly in internal VLA representations at inference time, keeping the input images and policy parameters unchanged.

\subsection{Mechanistic Interpretability and Representation Intervention.}
Sparse autoencoders (SAEs) decompose model activations into sparse features that can support mechanistic analysis~\cite{huben2024sparse,gao2025scaling}.
Their applications have extended from language models to visual concept analysis in CLIP~\cite{lim2025sparse,zaigrajew2025interpreting} and to cross-modal representation analysis in multimodal models~\cite{lou2025saev}.

Recent work has brought mechanistic analysis to VLA policies.
Directions associated with motion properties can steer robot behavior at inference time~\cite{haon2025mechanistic}.
SAE-based studies identify features associated with motion primitives, semantic concepts, and behavioral events, and test their influence through closed-loop interventions~\cite{swann2026sparse,jin2026event}.
These studies primarily examine nominal behavior, leaving the use of such features for adversarial defense underexplored.

Internal interventions have also served as defenses in vision and language models.
In CLIP, Dyslexify~\cite{hufe2025dyslexify} improves robustness to typographic attacks without fine-tuning by ablating attention heads that transmit typographic information to the CLS token.
Suppressing specific SAE features in CLIP's vision transformer also mitigates typographic attacks~\cite{joseph2025steering}.
Such interventions can also affect benign inputs.
In language models, amplifying refusal-related SAE features can improve resistance to jailbreaks but can also degrade performance on benign inputs~\cite{obrien2025steering}.
Some studies limit this degradation.
Adaptive activation steering defends vision-language models against jailbreaks from adversarial images by removing activation directions associated with harmful responses, adjusting the amount of removal for each input~\cite{wang2025steering}.
Conditional activation steering uses properties detected in hidden states to decide when to steer~\cite{lee2025programming}, and O'Brien et al.~\cite{obrien2025steering} show that conditioning SAE steering on a prompt-safety classifier mitigates the regressions on benign inputs.

Our work brings this conditional approach to adversarial patch defense in VLA models.
We suppress an SAE feature associated with patch presence only when a linear probe on the VLA's hidden state detects an attack.
We evaluate this intervention in closed-loop robot control and measure its effect on task success under both adversarial and benign conditions.

\section{Experimental Details}
\label{sec:experimental_details}

\paragraph{Models.}
We evaluate two VLA models, $\pi_{0.5}$~\cite{black2025pi05} and SmolVLA~\cite{shukor2025smolvla}, on the LIBERO-10 benchmark~\cite{liu2023libero}.
For $\pi_{0.5}$, we fine-tune the $\pi_{0.5}$-base checkpoint on the LIBERO training split following the official low-rank adaptation (LoRA) fine-tuning procedure.
To obtain a policy that uses only the agent camera, we replace all images from the wrist camera with black images during fine-tuning.
Fine-tuning runs for 30,000 steps on four NVIDIA RTX 6000 Ada GPUs (48 GB each).
For SmolVLA, we use the checkpoint released on Hugging Face without further fine-tuning.
The weights of both policies remain fixed in all defense experiments.

\begin{wrapfigure}{r}{0.48\textwidth}
    \vspace{-1.0em}
    \centering

    \setlength{\tabcolsep}{4pt}
    \begin{tabular}{ccc}
        \includegraphics[height=1.8cm]{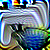}
         &
        \includegraphics[height=1.8cm]{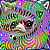}
         &
        \includegraphics[height=1.8cm]{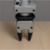}
        \\[-0.25em]

        {\scriptsize $\pi_{0.5}$}
         &
        {\scriptsize SmolVLA}
         &
        \\[-0.1em]

        \multicolumn{2}{c}{\small\textbf{(a)} UADA}
         &
        {\small\textbf{(b)} Gripper}
    \end{tabular}

    \vspace{-0.4em}
    \caption{Adversarial patches.}
    \label{fig:adversarial_patches}
    \vspace{-0.7em}
\end{wrapfigure}

\paragraph{Patches.}
Fig.~\ref{fig:adversarial_patches} shows the patches used in our experiments.
Our main attack is UADA~\cite{wang2025exploring}.
For each victim model, we optimize one universal patch on LIBERO-10 demonstration trajectories, following the original UADA implementation and optimization settings, including the patch size and geometric transformations.
The attack targets DoF 1 of the robot action.
Each optimized patch is shared across all LIBERO-10 tasks and evaluation rollouts and is applied to every camera view that the victim model uses.
For the additional evaluation in Appendix~\ref{sec:additional_experiments}, we also use a gripper patch, inspired by VLA-Fool~\cite{yan2025alignment}: an image of a robot gripper that we insert into the observation without optimizing it against the victim model.
In intermittent attacks, a patch is visible during $n\in\{3,10\}$ intervals of 10 consecutive frames each and absent otherwise.

\paragraph{SAE Training.}
Following the Dr. VLA implementation~\cite{swann2026sparse}, we train a separate TopK SAE~\cite{gao2025scaling} for each candidate layer.
The training data are representations from the residual stream, collected during clean LIBERO rollouts.
At each sampled control step, we mean-pool the representation over tokens and use the resulting vector as one training example.
As in Dr. VLA, we use a $1\times$ expansion ratio, so the SAE latent dimension equals the dimension of the VLA representation.
We use a smaller TopK sparsity than Dr. VLA, keeping $K=12$ active latents for $\pi_{0.5}$ and $K=6$ for SmolVLA.
We train each SAE for 6,677 optimization steps and follow Dr. VLA for all other architecture and optimization settings.

\paragraph{Attack-Related Feature Selection.}
We select the intervention target $(\ell^\star,j^\star)$ on 10 development rollouts, one per LIBERO-10 task, that are separate from the main evaluation.
In each rollout, the patch is inserted at randomly sampled intervals.
For each candidate layer-feature pair, we compute the Pearson correlation between the feature activation and the binary patch-presence indicator, separately for each rollout.
We then select the pair whose correlation exceeds $0.6$ in the largest number of rollouts and break ties by the mean correlation across rollouts.

\paragraph{Linear Probe Training.}
We train and evaluate probes on 100 rollout pairs, each consisting of a clean and an attacked rollout from the same initial state.
These rollouts are separate from those used for feature selection and from the 500 evaluation rollouts, and we split them 60/20/20 into training, development, and test sets within each task.
Frames without a visible patch are labeled negative and frames with a visible patch are labeled positive; each attacked rollout contains three 10-frame patch intervals.
For $\pi_{0.5}$, the probe is a class-balanced logistic regression on the token-mean-pooled prefix representation at layer 5 of the PaliGemma backbone (2048 dimensions), with the inverse regularization strength $C$ chosen from $\{0.01,0.1,1,10,100\}$.
For SmolVLA, the probe is a linear classifier trained with class-balanced binary cross-entropy and $L_2$ regularization on the mean over prefix tokens, excluding padding, at layer 9 of the VLM backbone.
For both models, we set the threshold $\tau$ on clean development data so that the FPR is at most $2\%$.
Under this constraint, we select SmolVLA's $C$ to maximize the detection rate on patch intervals of 10 frames in the development set.
We also train probes independently at other layers of the VLM and the vision encoder.

\paragraph{Intervention Strength.}
We apply Eq.~\eqref{eq:intervention} at the selected pair $(\ell^\star,j^\star)$ and set the strength $\lambda$ separately for each experimental condition.
We search $\lambda\in\{0.25,0.5,1,2,4\}$, adding $\{0.1,0.2,0.3\}$ in finer-grained experiments, select $\lambda$ on the development rollouts, and fix it before evaluation.
In the main evaluation, we use $\lambda=0.3$ for $n=3$ and $\lambda=0.2$ for $n=10$ on $\pi_{0.5}$, and $\lambda=0.2$ for $n=3$ and $\lambda=0.25$ for $n=10$ on SmolVLA.
For SmolVLA, we use $\lambda=0.25$ under benign conditions and $\lambda=0.1$ under persistent attacks.
Fig.~\ref{fig:intervention_strength} reports the sensitivity of SR to $\lambda$ on $\pi_{0.5}$ under both attack and benign conditions.
Although we tune $\lambda$ separately for each attack frequency, the selected values remain within a narrow range ($0.2$--$0.3$) across $n=3$ and $n=10$ for both models.
This suggests that, in practice, a single intervention strength selected on representative development rollouts could be fixed during inference without prior knowledge of the attack frequency.

\section{Additional Experiments}
\label{sec:additional_experiments}

We further analyze our method beyond the main evaluation.
We evaluate
(1) performance under gripper patch attacks,
(2) detector specificity across patch types,
(3) the effect of attack-related feature selection, and
(4) performance under persistent attacks.

\subsection{Evaluation under Gripper Patch Attacks}

\begin{wraptable}{r}{0.40\textwidth}
    \vspace{-4em}
    \centering

    \caption{SR (\%) under gripper patch attacks.}
    \label{tab:gripper_patch_results}

    \small
    \setlength{\tabcolsep}{5pt}
    \renewcommand{\arraystretch}{0.88}

    \begin{tabular}{lcc}
        \toprule
        \multirow{2}{*}{Method}
         & \multicolumn{2}{c}{\textbf{Victim: $\pi_{0.5}$}}                 \\
        \cmidrule(lr){2-3}
         & $n=3$                                            & $n=10$        \\
        \midrule

        Baseline
         & 30.4                                             & 4.4           \\

        \textbf{Ours}
         & \textbf{46.4}                                    & \textbf{33.6} \\

        \midrule

        w/o Det.
         & 0.0                                              & 0.0           \\

        w/ Oracle Det.
         & 50.0                                             & 31.8          \\

        \bottomrule
    \end{tabular}

    \vspace{-0.5em}
\end{wraptable}

We evaluate $\pi_{0.5}$ under gripper patch attacks, in which the inserted patch is an unoptimized image of a robot gripper (Fig.~\ref{fig:adversarial_patches}(b)).
For this setting, we train the probe on rollouts with the gripper patch.
The unoptimized gripper patch lowers baseline SR more than UADA does: to $30.4\%$ for $n=3$ and $4.4\%$ for $n=10$, compared with $46.8\%$ and $26.0\%$ under UADA (Tables~\ref{tab:gripper_patch_results} and~\ref{tab:main_results}).
Our method raises SR to $46.4\%$ and $33.6\%$, respectively, which is comparable to oracle detection ($50.0\%$ and $31.8\%$).
Intervention without detection reduces SR to $0\%$ in both settings.
As with UADA, controlling when to intervene remains important under this patch type.

\subsection{Detector Specificity Across Patch Types}

\begin{wraptable}{r}{0.48\textwidth}
    \vspace{-4.0em}
    \centering

    \caption{Detection performance (\%) across patch types.}
    \label{tab:anomaly_detection}

    \vspace{0.1em}

    \small
    \setlength{\tabcolsep}{3.5pt}
    \renewcommand{\arraystretch}{0.92}

    \resizebox{\linewidth}{!}{%
        \begin{tabular}{llccc}
            \toprule
            Training Patch & Test Patch & Recall          & FPR           & Precision      \\
            \midrule

            \multirow{4}{*}{UADA}
                           & UADA       & \textbf{100.00} & 1.47          & \textbf{99.62} \\
                           & White      & 5.74            & 1.47          & 79.59          \\
                           & Black      & 2.17            & 1.47          & 59.56          \\
                           & Noise      & \textbf{100.00} & 1.47          & 98.55          \\

            \midrule

            \multirow{4}{*}{Gripper}
                           & Gripper    & \textbf{100.00} & 9.45          & \textbf{99.62} \\
                           & White      & 23.11           & 9.45          & 70.97          \\
                           & Black      & 16.32           & 9.45          & 63.33          \\
                           & Noise      & 19.71           & 9.45          & 67.59          \\

            \bottomrule
        \end{tabular}%
    }

    \vspace{-0.5em}
\end{wraptable}

To measure how specific the probe is to its training patch type, we evaluate it on its training patch as well as on white, black, and random-noise patches (Table~\ref{tab:anomaly_detection}).
Both probes achieve $100\%$ recall on their respective training patches.
The probe trained on UADA patches rarely flags white or black patches (recall $5.74\%$ and $2.17\%$) but flags every frame with a random-noise patch (recall $100\%$).
The UADA probe therefore detects random noise as well as UADA patches.
The probe trained on gripper patches detects $16.32$--$23.11\%$ of white, black, and noise patches, with a higher FPR of $9.45\%$ compared to $1.47\%$ for the UADA probe.
These differences suggest that detector specificity depends on the patches used to train the probe.

\subsection{Effect of Attack-Related Feature Selection}

\begin{wrapfigure}{r}{0.42\textwidth}
    \vspace{-3.5em}
    \centering
    \includegraphics[width=\linewidth]{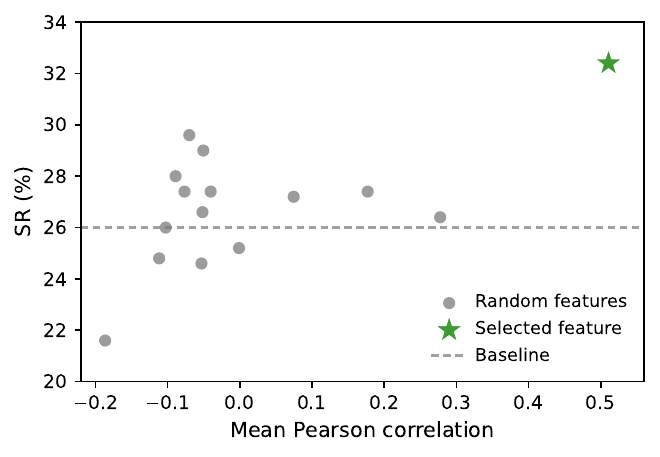}
    \caption{Correlation with patch presence and SR of the selected and random features ($\pi_{0.5}$, $n=10$).}
    \label{fig:feature_correlation}
    \vspace{-0.8em}
\end{wrapfigure}

We test how the choice of the suppressed feature affects SR.
For $\pi_{0.5}$ at $n=10$, we compare the selected feature with SAE features sampled randomly from layers deeper than the detection layer, with identical detection and intervention settings.
For each feature, we measure its mean Pearson correlation with patch presence and the SR obtained by suppressing it.
As shown in Fig.~\ref{fig:feature_correlation}, the selected feature has the highest correlation and yields the highest SR.
Some random features also improve SR over the baseline, but by smaller margins, and others reduce it.
These results support selecting features by their correlation with patch presence.

\needspace{10\baselineskip}
\subsection{Performance under Persistent Attacks}

\begin{wraptable}{r}{0.40\textwidth}
    \vspace{-2.0em}
    \centering
    \caption{SR (\%) under persistent attacks.}
    \label{tab:persistent_attack_results}
    \small
    \setlength{\tabcolsep}{7pt}
    \renewcommand{\arraystretch}{0.90}

    \begin{tabular}{lcc}
        \toprule
        Method
         & $\pi_{0.5}$
         & SmolVLA                                            \\
        \midrule

        Baseline
         & 12.4
         & 1.8                                                \\

        \textbf{Ours}
         & 9.0 {\scriptsize\textcolor{red}{$\downarrow 3.4$}}
         & 1.6 {\scriptsize\textcolor{red}{$\downarrow 0.2$}} \\

        \midrule

        w/o Det.
         & 7.4 {\scriptsize\textcolor{red}{$\downarrow 5.0$}}
         & 1.6 {\scriptsize\textcolor{red}{$\downarrow 0.2$}} \\

        w/ Oracle Det.
         & 7.4 {\scriptsize\textcolor{red}{$\downarrow 5.0$}}
         & 1.6 {\scriptsize\textcolor{red}{$\downarrow 0.2$}} \\

        \bottomrule
    \end{tabular}

    \vspace{-0.5em}
\end{wraptable}

We also evaluate a persistent attack in which the patch remains visible throughout the rollout.
Because the patch appears in every frame, oracle detection is equivalent to continuous intervention in this setting.
As shown in Table~\ref{tab:persistent_attack_results}, no intervention method improves SR.
For $\pi_{0.5}$, our method lowers SR from $12.4\%$ to $9.0\%$, and both \textit{w/o Det.} and \textit{w/ Oracle Det.} lower it to $7.4\%$.
For SmolVLA, all three intervention methods yield $1.6\%$, compared with $1.8\%$ for the baseline.
Even with oracle detection, suppressing the selected feature does not recover SR under persistent attacks, which is a limitation of our approach.

\end{document}